# Rumor Detection with A Novel Graph Neural Network Approach

Tianrui Liu[1, *], Qi Cai[2], Changxin Xu[3], Bo Hong[4], Fanghao Ni[5], Yuxin Qiao[6], Tsungwei Yang[7]

[1] Electrical and Computer Engineering, University of California San Diego, La Jolla, USA
[2] Computer Science and Engineering, University of North Texas, Denton, USA
[3] Computer Information Technology, Northern Arizona University, Flagstaff, USA
[4] Computer Information Technology, Northern Arizona University, Flagstaff, USA
[5] Data Science and Artificial Intelligence, Indian Institute of Technology Guwahati, Assam, India
[6] Computer Information Technology, Northern Arizona University, Flagstaff, USA
[7] Computer Science, Tunghai University, Taichung, Taiwan

* Corresponding author: Tianrui Liu (Email: tianrui.liu.ml@gmail.com)

**Abstract:** The wide spread of rumors[3] on social media has caused a negative impact on people's daily life, leading to potential panic, fear and mental health problems for the public.[47] How to debunk rumors as early as possible remains a challenging problem. Existing studies mainly leverage information propagation structure to detect rumors[6], while very few works focus on correlation among users that they may coordinate to spread rumors in order to gain a large popularity. In this paper, we propose a new detection model, that jointly learns both the representations of user correlation and information propagation to detect rumors on social media. Specifically, we leverage graph neural networks to learn the representations of user correlation from a bipartite graph[5] that describes the correlations between users and source tweets[12], and the representations of information propagation with a tree structure. Then we combine the learned representations from these two modules to classify the rumors. Since malicious users intend to subvert our model after deployment, we further develop a greedy attack scheme to analyze the cost of three adversarial attacks: graph attack, comment attack and joint attack. Evaluation results on two public datasets illustrate that the proposed MODEL outperforms the state-of-the-art rumor detection models. We also demonstrate our method performs well for early rumor detection.[8] Moreover, the proposed detection method is more robust to adversarial attacks compared to the best existing method. Importantly, we show that it requires high cost for attackers to subvert user correlation pattern, demonstrating the importance of considering user correlation for rumor detection.

## 1. Introduction

This paper proposes a novel rumor detection model, which jointly learns the representations of user correlation and information propagation to detect rumors on social media. Nowadays, rumors are like social virus that spread fast, deep and broadly across social media, such as Twitter and Facebook, which has caused potential threat and panic to people's daily life. For example, a large number of rumors about COVID-19 pandemic make a ton of people feel uneasy and scared in the year of 2020. Hence, how to effectively detect rumors[2] and prevent them from diffusion as early as possible has attracted a lot of attention from both industry and academia in recent years.

The early works mainly adopt traditional machine learning methods[59] based on handcrafted features, such as user profile and text contents, to detect rumors online. However, the traditional approaches, such as SVM, Decision Tree and Random Forest, heavily depend on feature engineering[53], which is time-consuming and can be easily manipulated by malicious users. Recent studies developed deep learning models, such as bidirectional transformers(BERT)[10](which are widely using in review tasks[13]), recurrent neural network(RNN),[27] recursive neural networks (RvNN)[39], Long-Short Term Memory (LSTM), generative adversarial network (GAN)[16,56], transformer[7,20] and Convolutional Neural Networks (CNN)[11,17], to learn sequential features from information propagation patterns over time.[1] These models also have widespread applications in other fields, such as data security [21,22], vision learning [18,52], material analysis [15,51] , compiling [33] and hardware designing[34], E-commerce[37], image segmentation[43], traffic controlling [38], communication [28,32] and Aerial Search [30] . These methods, however, only learn the correlations from local neighbors in the structure of information propagation while ignore the global structures of rumor dispersion.[32] In order to better learn the representations of misinformation propagation, researchers recently developed a graph convolutional neural networks (GCN)[40] based detection model, BiGCN, that combines top-down with bottom-up tree propagation structure to deal with the propagation and dispersion of rumors. While BiGCN improves the detection accuracy, it does not take user behaviors of misinformation campaigns into account. In practice, some malicious users coordinate to report some same claims together in order to make rumors spread faster and more broadly within a short time across social media. This global user dependency relationship is helpful to find the correlations among different users and source tweets. It thus can improve the detection accuracy of the coordinated misinformation campaigns.[14]

Inspired by this insight, we propose a joint detection model, called MODEL, which encodes both user correlation pattern and information propagation pattern to improve the accuracy of rumor detection. The proposed method learns the representations from a user-tweet bipartite graph and from a tree structure of information diffusion, as shown in Fig. The user-tweet bipartite graph, consisting of users and source tweets, denotes users perform actions, such as post, retweet and comment, on the source tweets.[46]The tree structure of information propagation shows the information of a source tweet forwards from a root node (source tweet) to its children's nodes (comments/retweet) in a tree.[29] After that, we combine the learned representations from these two parts and then feed them into a fully connected layer[55] to classify the source tweets into different categories. In order to enhance the influence of root tweets in a deep tree, we use the same methodology as BiGCN to pass the features of root tweets to their children's nodes.[44]

The main contributions of this paper are summarized as follows:

We propose a new detection model, which learns the joint representations of user correlation and information propagation to detect rumors on social media.

We adopt a greedy attack scheme that generates fake edges and comments to explore the cost of adversarial attacks on the proposed method.

Evaluations on two public datasets demonstrate the proposed model outperforms the baselines in terms of detection accuracy and F1 score. We also show attackers need higher cost to fool our method compared to the best baseline.

The rest of paper is organized as follows. Section 1 introduces the problem of rumor detection and the background of graph neural networks. Then we present the overall architecture of the proposed model in Section 2 and 3 In Section 4, we evaluate the performance of the proposed model on two public datasets. Finally, we conclude the paper in Section 5.

## 2. Organization of the Text

We first introduce the problem of rumor detection, and then present the background of graph convolutional neural networks used in the proposed model.

### 2.1. Problem Statement

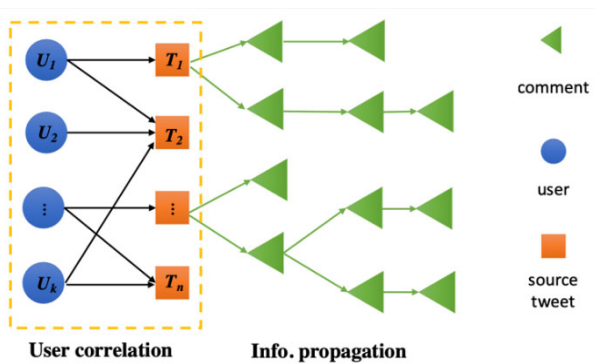

**Figure 1.** A toy example of rumor detection

Fig.1 illustrates an example of our rumor detection method, consisting of two parts: user correlation and information propagation. For user correlation, let $\{T_1, T_2, ... T_n\}$ denote a set of $n$ source claims (tweets) made by many different users on social media, $U_1, U_2, ... U_k$ denote a group of $k$ users who perform actions, such as post and retweet, on source tweets. In this paper, a bipartite graph, $G_1 = (V, E, A_g)$ is constructed to describe the user-tweet relationship, where $V$ denotes the vertices, and $E$ denotes the edges. When a user $U_j$ performs an action on a source tweet $T_i$, there is an edge between them. Mathematically, the relationship between users and source tweets can be denoted by an adjacency matrix. For information propagation module, each source tweet $T_i$ consists of $m$ comments with a tree structure of information diffusion, denoted by $\{C_1, C_2, \cdots, C_m\}$. We use $G_2 = (V, E, A_t)$ to denote the tree structure. Similar to user-tweet relationship, we adopt an adjacency matrix, $A_t$, to denote the relationship among (root) source tweets and the corresponding comments in a tree structure.

The goal of this paper is to learn the representation of source tweets from both user correlation and information propagation, and then classify the source tweets into $d$ different categories based on the learned representations.

### 2.2. Graph Convolutional Neural Network

Graph neural networks (GNNs) aim to encode[23] structural information about a graph into a low-dimensional latent space, which generally consists of two components 1) *graph convolution layers* which extract substructure features from local nodes; 2) *a graph aggregation layer* that aggregates node-level features into a graph-level representation. As one of the most popular GNN, graph convolutional neural networks (GCNs) have been widely used in a variety of applications[36], such as classification and clustering,[54] because they can encode both node features and graph structure of neighboring nodes. The basic idea of GCNs is to learn the representations of nodes in a graph based on the input features of nodes, $X$, and their graph structure, denoted by an adjacency matrix $A$. While there are many different variants of GCNs in the literature, the general form of a GCN is expressed by:

$$H^{l+1} = \sigma(\hat{A}H^{(l)}W^{(l)}),$$

Where $\hat{A} = \widetilde{D}^{-\frac{1}{2}}\widetilde{A}\widetilde{D}^{-\frac{1}{2}}$ is the normalized adjacency matrix. In addition, $W^l$ is the weight matrix to be trained, and $\sigma(\cdot)$ is the activation function, such as ReLU.[41]

## 3. System Design and Modeling

In this section, we introduce the overall architecture of the proposed our model for rumor detection. The core idea of our model is to learn the representations of source tweets from both user correlation and information propagation respectively, which are then adopted to classify rumors. The reason why we learn these two representations independently is that the embeddings of user features and comments may be not in the same latent space. The components of the model are discussed below:

### 3.1. User feature encoding

To reduce account maintenance cost, many user accounts would be created within a short period of time. Inspired by this, we attempt to utilize user profiles as features to depict how users participate in rumor dispersion.[4] The extracted features include number of followers, number of friends, account created time, the total number of tweets and so on. We feed user features into fully-connected layers to learn its embedding for further optimization[45]

### 3.2. Text encoding

In order to capture the sequential information present in source tweets and the corresponding responses such as comments, this paper employs the Gated Recurrent Unit (GRU) as an encoder to derive a feature vector.[9] Given the variance in sequence length of tweets, zero padding is performed to standardize the length of each sequence. The text data from tweets is then embedded into a lower dimension,[50] and pre-trained word embeddings specific to Twitter data, provided by the GloVe algorithm, are utilized. [24] These embeddings are then fed into the GRU to learn the representation of the Twitter text content. This approach facilitates a nuanced understanding of the sequential nature of the text data, which is instrumental in analyzing and detecting rumors on social media platforms.[25]

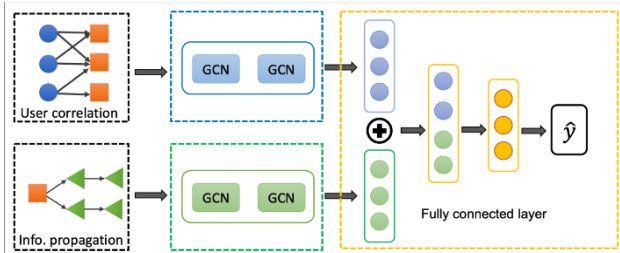

**Figure 2.** Our proposed model architecture

### 3.3. User Correlation Representation

In addressing the scenario where a collective of users may collaborate to propagate false claims swiftly and extensively across social media, this paper delves into exploring correlations among different users and their claims. To achieve this, a user correlation pattern is utilized to learn the representations of source tweets within a user-tweet graph. A bipartite graph is constructed to delineate the relationship between users and tweets, which then serves as a foundation for applying a two-layer Graph Convolutional Neural Networks (GCN) to learn the representations of source tweets.

To facilitate the scalability of the algorithm, especially in the context[19] of large-scale graphs encompassing hundreds of thousands of nodes, a subgraph approach is adopted. By focusing on a subgraph with k-hop neighboring nodes of source tweets as input for the GCN model, the training time for graph-based rumor detection is significantly reduced[57], making the methodology more efficient and applicable to large-scale scenarios.[35] This streamlined approach aims to enhance the practicality and effectiveness of the rumor detection process, especially in a large-scale, real-world social media environment.[31]

### 3.4. Information Propagation Representation

In order to understand the flow of source tweets to their respective child nodes, a deep learning model rooted in a tree propagation structure is devised to learn their representations. The methodology employed draws parallels to a referenced approach for constructing a top-down tree concerning a source tweet and its ensuing comments. The merit of utilizing a top-down tree lies in its capability to aptly capture the causal features intrinsic to a rumor propagation thread. Analogous to the approach taken for user correlation representation, a Graph Convolutional Neural Networks (GCN) with a directional graph is adopted to learn the representations of information propagation. This method is geared towards effectively understanding the dynamics of rumor spread and the interactions within a thread, which is crucial for accurate rumor detection.[42]

### 3.5. Rumor Classification with Joint Representations

The methodology employs the combined representations of user correlation and information propagation to facilitate rumor detection. Initially, the representations concerning user correlation and information propagation are concatenated. Following this step, the merged representations are fed into fully connected layers, culminating in a softmax layer.[49] This structure aids in categorizing the rumors into distinct classifications. By harnessing both the interactions among users and the dissemination patterns of information, this procedure aspires to enhance the accuracy of rumor detection across social media platforms.[60]

### 3.6. Model training

The proposed model is trained by minimizing the Cross-Entropy loss between the predictions and the ground truth. During training, the model parameters are updated using back propagation until the loss converges or the maximum iterations are met.[48] To summarize, in this system, we first compute the representations of source tweets from both user correlation and information propagation, and then feed them into fully connected layers to classify the source tweets.

## 4. Evaluation

In this section, extensive experiments are carried out to evaluate the performance of the proposed model on two benchmark datasets, Twitter15 and Twitter 16. First of all, we compare the detection accuracy of our method with the baseline methods. In order to demonstrate the robustness of our method, we further analyze the cost of adversarial attack. Also, we demonstrate the good performance of the proposed method for early rumor detection.[58] Finally, we conduct ablation studies to compare MODEL with its variants.

---

**Algorithm 1** The proposed REPORT algorithm.

1: **Input:** adjacency matrix $A_g$ and $A_t$ for user-tweet graph and tree structure, user features $X_u$, tweet embeddings, $E_t$, source tweets, $\mathcal{T}$.
2: **Output:** Rumor classification result, $\hat{y}$.
3: **for** batch size of $\mathcal{T}$ **do**
4:   Encode user features, $H_u$, using Eq. (2).
5:   Encode tweet text, $S_t$, using Eq. (3).
6:   Learn representations of user correlation, $H_g$, using Eq. (4).
7:   Learn representations of information propagation, $\hat{H}_t$, using Eq. (7).
8:   Joint representations of source tweets, $H$, using Eq. (8).
9:   Predict results, $\hat{y} = softmax(FC(H))$.
10:  **Return** $\hat{y}$.
11: **end for**

---

### 4.1. Dataset

To evaluate the performance of the proposed MODEL, we conduct extensive experiments on two public Twitter datasets, Twitter15 and Twitter16. These two datasets contain four labels: Non-rumor (N), False Rumor (F), True Rumor (T), and Unverified Rumor (U). To construct a completely connected user-tweet graph, we choose users that report at least two source tweets or responsive post. Since user profiles are not available in the original datasets, we collect them from Twitter API 1. Table 1 illustrates the statistics of the two benchmark datasets.[61] Note that, the total number of source tweets in this paper is less than that in the original datasets, because some source tweets and users are suspended at the time of collecting data[26] from Twitter this year. Like the previous studies, we split the datasets into training data and testing data with 80% and 20%, respectively. In addition, to better evaluate the performance of our method, we adopt $k$-fold ($k = 5$) cross-validation method to conduct experiments in this paper.

Table 1. Twitter datasets statistics

| Statistics | Twitter15 | Twitter16 |
|---|---|---|
| # of Source | 1473 | 790 |
| # of User applied | 47925 | 24047 |
| # of True rumors | 366 | 203 |
| # of False rumors | 360 | 192 |
| # of Unverified | 372 | 196 |
| # of Non-rumors | 374 | 199 |

Table 2. Experiment results

| Methods | Twitter15 | | | | | Twitter16 | | | | |
|---|---|---|---|---|---|---|---|---|---|---|
| | Accuracy | N (F1) | F (F1) | T (F1) | U (F1) | Accuracy | N (F1) | F (F1) | T (F1) | U (F1) |
| CEM | 0.436 | 0.394 | 0.459 | 0.512 | 0.410 | 0.415 | 0.465 | 0.394 | 0.354 | 0.454 |
| DTC | 0.412 | 0.394 | 0.628 | 0.296 | 0.269 | 0.392 | 0.429 | 0.450 | 0.301 | 0.378 |
| RvNN | 0.723 | 0.682 | 0.758 | 0.821 | 0.654 | 0.737 | 0.662 | 0.743 | 0.835 | 0.708 |
| SVM-TK | 0.679 | 0.707 | 0.677 | 0.715 | 0.777 | 0.665 | 0.613 | 0.658 | 0.777 | 0.780 |
| BiGCN | 0.827 | 0.805 | **0.829** | **0.887** | 0.788 | 0.821 | 0.747 | 0.802 | **0.907** | 0.829 |
| REPORT | **0.860** | **0.938** | 0.828 | 0.877 | **0.797** | **0.853** | **0.841** | **0.831** | 0.903 | **0.835** |

## 4.2. Experiment results

We first compare the detection performance of the proposed MODEL with baselines above on Twitter15 dataset. We use accuracy and F1 score as the evaluation metrics in the experiments. Table 2 illustrates the comparison results of different rumor detection models. It can be observed that our method outperforms the baseline methods in term of accuracy and F1 score. The reason why our method beats BiGCN is that we can learn the correlations among different sources tweets with user correlation pattern while the latter does not. Thus, we can see the effectiveness of incorporating user correlation pattern for rumor detection. For Twitter16 dataset, we also show that the proposed method has a better detection performance than the baselines, as shown in Table 2.

## 5. Conclusion

Rumor detection algorithms are essential for identifying and curbing the spread of misinformation online. They ensure that only verified information circulates, safeguarding public safety, protecting financial markets, and maintaining national security. Advanced models, especially those using graph neural networks, enhance early detection and accuracy, mitigating the potential negative impacts of false information swiftly.

In this paper, we developed a novel rumor detection model, that jointly learns the representations of user correlation and information propagation based on graph neural networks. Specifically, it first leverages graph convolutional neural networks (GCNs) to learn the representations of source tweets from a user-tweet bipartite graph and a tree propagation structure, respectively. Then we combine the learned representations to classify rumors. Evaluations on two publicly available datasets show that our model has the best detection performance compared to the existing detection models. It also has a higher detection accuracy for early rumor detection compared to the baselines.

In the future, we would like to investigate more powerful approaches to learn representation of text information, which is a viable way to gain a better model performance.